\documentclass[10pt,twocolumn,letterpaper]{article}

\usepackage[square,numbers]{natbib}
\usepackage{iccv}
\usepackage{times}
\usepackage{epsfig}
\usepackage{graphicx}
\usepackage{amsmath}
\usepackage{amssymb}
\usepackage{amssymb}
\usepackage{color}

\usepackage[breaklinks=true,bookmarks=false]{hyperref}

\iccvfinalcopy % *** Uncomment this line for the final submission

\def\iccvPaperID{****} % *** Enter the ICCV Paper ID here

\begin{document}

%%%%%%%%% TITLE
\title{Predictive Coding Networks Meet Action Recognition}
%Motion Feature Generating Networks for Action Recognition
%Predictive Network Meets Action Recognition
\author{Xia Huang$^1$ \;\;\;  Hossein Mousavi$^1$ \;\;\; Gemma Roig$^{1,2}$\\
$^1$Singapore University of Technology and Design, Singapore\\
$^2$Department of Computer Science, Goethe University of Frankfurt am Main, Frankfurt am Main, Germany\\
{\tt\small xia\_huang@mymail.sutd.edu.sg \;\; hossein\_mousavi@sutd.edu.sg \;\; gemmar@mit.edu}
}

\maketitle
%\thispagestyle{empty}

%%%%%%%%% ABSTRACT
\begin{abstract}
Action recognition is a key problem in computer vision that labels videos with a set of predefined actions. Capturing both, semantic content and motion, along the video frames is key to achieve high accuracy performance on this task. Most of the state-of-the-art methods rely on RGB frames for extracting the semantics and pre-computed optical flow fields as a motion cue. Then, both are combined using deep neural networks. Yet, it has been argued that such models are not able to leverage the motion information extracted from the optical flow, but instead the optical flow allows for better recognition of people and objects in the video. This urges the need to explore different cues or models that can extract motion in a more informative fashion. To tackle this issue, we propose to explore the predictive coding network, so called PredNet, a recurrent neural network that propagates predictive coding errors across layers and time steps. We analyze whether PredNet can better capture motions in videos by estimating over time the representations extracted from pre-trained networks for action recognition. In this way, the model only relies on the video frames, and does not need pre-processed optical flows as input. We report the effectiveness of our proposed model on UCF101 and HMDB51 datasets.     

\end{abstract}

%%%%%%%%% BODY TEXT
\section{Introduction}

\begin{figure}[t!]
\begin{center}
   \includegraphics[width=1\linewidth]{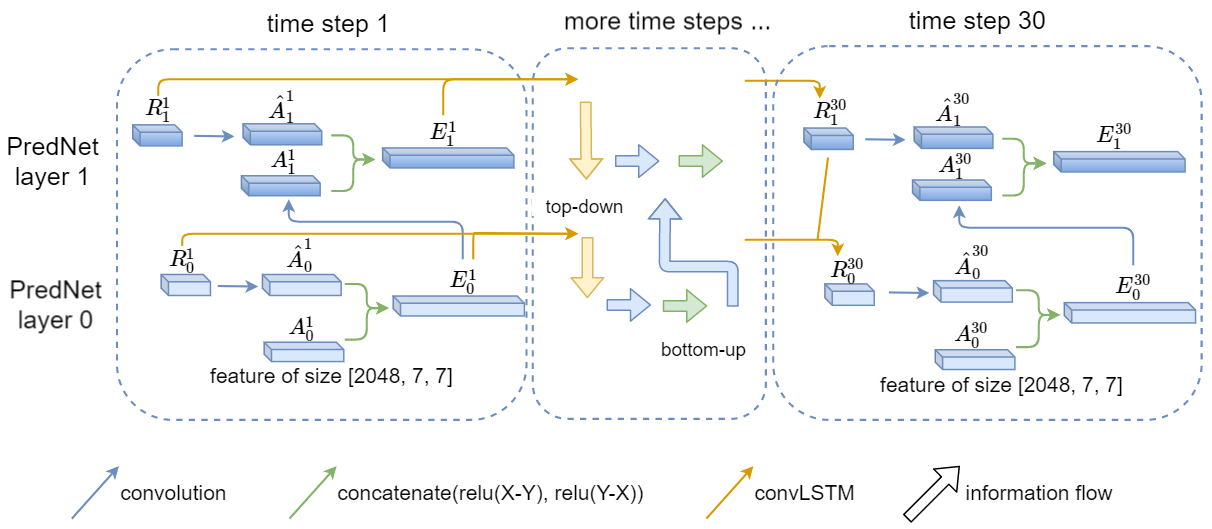}
\end{center}
   \caption{PredNet architecture. There are four main components in each time step -- R, the representation unit, \^{A}, the input unit, A, the prediction unit and E, the error unit. The subscript annotates the time step while the superscript indicates which layer the unit is in. Starting from time step 2, the model follows a top-down then bottom-up process then move to time step 3 and more. We combine PredNet with ResNet-50 as our model. There are a total of 30 time steps in our layout.}
\label{fig:prednet}
\end{figure}

Video-based understanding tasks are important computer vision problems which have many applications, such as smart surveillance, human-machine interaction, human behaviour understanding and online video retrieval.
Action classification~\cite{he2016deep} is at the core of most of the aforementioned applications, and has been extensively studied~\cite{kuehne2011hmdb,soomro2012ucf101, kay2017kinetics, karpathy2014large, simonyan2014two, carreira2017quo,hara2018can, monfort2019moments}. 

With the rise of deep learning models for object recognition~\cite{krizhevsky2012imagenet, simonyan2014very, he2016deep}, such models have also been successfully applied to action recognition~\cite{tran2014learning, yue2015beyond, tran2015learning, simonyan2014two, xian2018feature, carreira2017quo, zhu2018end, wang2016temporal,ravanbakhsh2015action}, after large-scale datasets of videos with action labels collected were made available online~\cite{monfort2019moments, kay2017kinetics}.

Compared to recognition in still images,  motion and the temporal component of videos play an important role for action recognition. For this reason, most of the current state-of-the-art models, pre-process the videos to extract optical flow fields between contiguous video frames, and use models that can ingest sequences, either with recurrent neural networks or LSTM~\cite{sutskever2014sequence, hochreiter1997long, pearlmutter1989learning}, or with feedforward networks that look at the entire video sequence at once~\cite{simonyan2014two, carreira2017quo, zhu2018end, wang2016temporal}. It has been recently argued that the pre-processed optical flow does not bring motion information into the models~\cite{sevilla2018integration}. Instead, it brings more semantic information as the optical flows can also be interpreted as object masks. Two of the observations in the experiments in~\cite{sevilla2018integration} are that 1) optical flow is invariant to the appearance such that models can recognize actions without the assist of the colors/appearance of the objects, 2) the small movements and boundary accuracy in optical flow are most correlated to action recognition performance. Hence, they argued that other motion cues should be explored and incorporated to the models.

Besides, it is known that optical flow is computationally expensive and time consuming~\cite{revaud2015epicflow, weinzaepfel2013deepflow, bao2014fast, brox2010large}. The ground-breaking models~\cite{dosovitskiy2015flownet, ilg2017flownet} constructed  CNNs to solve this problem as a supervised learning task but still relied heavily on large dataset for pre-training to achieve good performance.

In this paper, we aim to explore a neural network architecture for extracting motion information directly from the video frames. We propose to use predictive coding neural networks, also called PredNet~\cite{lotter2016deep}, displayed in Figure~\ref{fig:prednet}. PredNets have been introduced for predicting the next frame in a video, at pixel level, which allows training the model on large sets of unlabeled videos. The learned representation can then be used for another video task, such as predicting the direction of a car steering wheel angle, as described by~\cite{lotter2016deep}. We argue that PredNet, with its bottom-up and top-down deep recurrent connections, resembles the way optical flow is produced but more informative as this pixel movement is captured in the ''PredNet process''. 

Inspired by the above observations and PredNet, we propose a new model which is the combination of the CNN model ResNet-50~\cite{he2016deep} and PredNet without directly using optical flow for action recognition. ResNet-50 processes the RGB frames while PredNet takes the feature extracted from ResNet-50 and predicts the feature in the next time step. Then, the extracted features from ResNet-50 and PredNet are concatenated to fed them to the action prediction classifier. 

We demonstrate the power of PredNet for action recognition in HMDB51~\cite{kuehne2011hmdb} and UCF101~\cite{soomro2012ucf101}. Even though we do not use optical flow, we achieve competitive results compared to state-of-the-art models that implicitly use optical flow fields pre-computed from the video frames as input to their model.

\subsection{Related work}
Action recognition have been extensively investigated in the past  years. Here, we briefly discuss some of the methods, distinguishing into  before and after deep neural networks models were introduced for action recognition. 

\paragraph{Hand-crafted features based Models} Most of the early proposed methods for action recognition relied on extracting local hand-crafted features from videos~\cite{dollar2005behavior,wang2009evaluation,willems2008efficient}. A plethora of features were proposed, such as Histogram of Gradients (HOG)~\cite{dalal2005histograms}, Histogram of Gradient in 3D including the time component (HOG3D)~\cite{klaser2008spatio},  Motion Boundary Histogram~\cite{dalal2006human}, and Histogram of OpticalFlow (HOF)~\cite{laptev2008learning}. The video features were extracted densely or around space time interest points, for example Harris3D~\cite{laptev2005space}. Then, such local features were encoded into video features using bag of features (BoF)~\cite{o2011introduction}, also including the spatio-temporal coordinates~\cite{sheikh2005exploring}. Others, used the local features to track the moving objects along the video frames~\cite{peng2016bag,peng2014action}. Once the video feature were computed, a classifier such as a support vector machine, was used to predict the action label. All of these methods are now surpassed by deep learning models, which we describe next\cite{krizhevsky2012imagenet,simonyan2014very,he2016deep,szegedy2015going}.

\paragraph{Deep learning models} With the success of deep learning models in many computer vision tasks, those have also been explored for action recognition, and are the current state-of-the-art models for this task. Several models have been proposed in the literature that differ mainly in the architecture structure and the input used. Some examples are convolutional neural networks with LSTM~\cite{donahue2015long, yue2015beyond}, two-stream networks\cite{feichtenhofer2016convolutional, simonyan2014two} which use optical flow as input, 3D ConvNet\cite{taylor2010convolutional, ji20123d}. Combinations of those, such as a two-stream 3D ConvNet have also been explored~\cite{carreira2017quo}, as well the addition of  attention mechanisms~\cite{sharma2015action, song2017end,vaswani2017attention}. Other major differences of these models lie in the way they process temporal information, by average/max pooling\cite{zhu2018end}, and some models have explored adding  skeleton data~\cite{song2017end}.

\section{Approach}

We use a pre-trained CNN model~\cite{he2016deep} for extracting features at frame level, and then a PredNet~\cite{lotter2016deep} to predict the  feature representation of the next video frame. We merge all those representations to predict the action class label. Below we describe each of the components of our model. 

\begin{figure*}[t]
\begin{center}
   \includegraphics[width=1\linewidth]{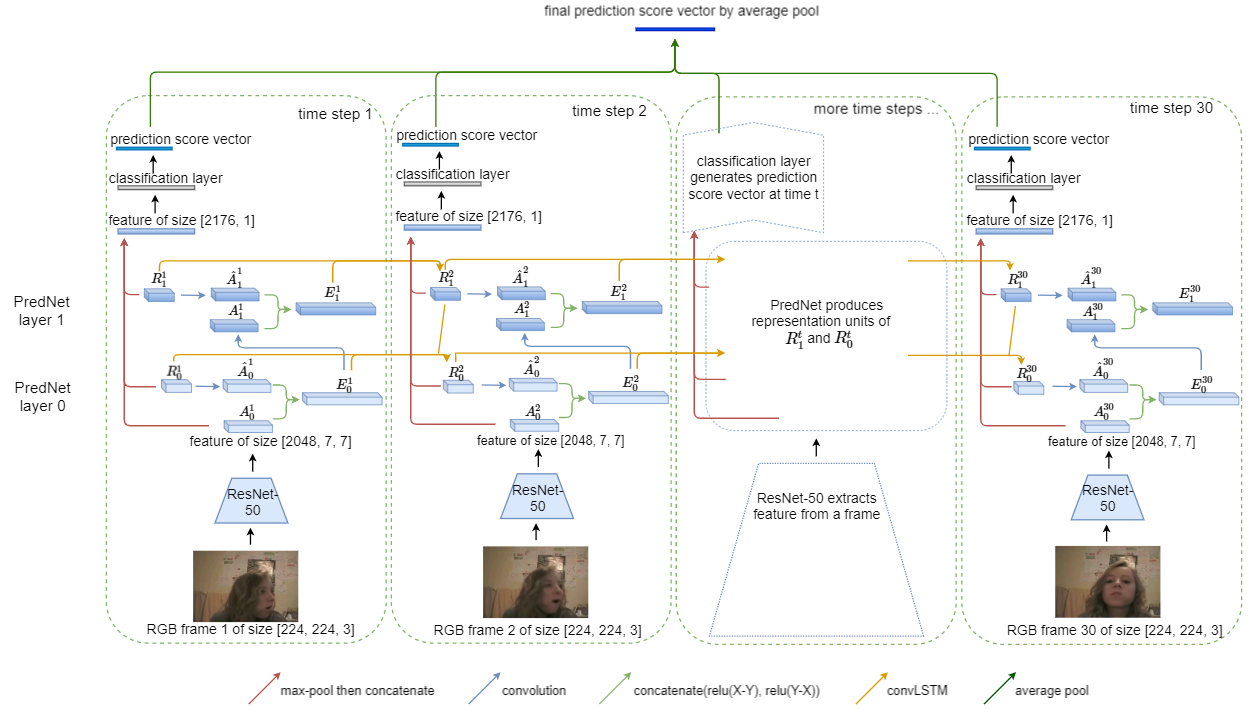}
\end{center}
   \caption{ResNet-50 + PredNet structure. At time step t, the corresponding frame is sent to a pre-trained  ResNet-50 to extract features, which becomes PredNet's input (A$^{t}_{0}$ unit). After the process is finished at that time step in PredNet (described in Figure 1 and Predictive Network), we apply max-pool on A$^{t}_{0}$, R$^{t}_{0}$ and R$^{t}_{1}$ then concatenate them as one feature for classification. The prediction score vector comes out from the classification layer. After the total number of time steps, average pool is applied on all of the prediction score vectors to generate the final one prediction score vector. In our model, PredNet has 2 layers and there are 30 time steps in total.}
\label{fig:onecol}
\end{figure*}

%what is the method -- what are the specifications for PN and RN.
\paragraph{Semantic features extracted from a CNN} We use a pre-trained CNN to extract features at the frame level. The importance of using pre-trained models on large datasets, for boosting the classification accuracy in action recognition has also  been emphasized by others~\cite{hara2018can}~\cite{zisserman14a}~\cite{monfort2019moments}. We use a ResNet-50~\cite{he2016deep}, which consists of 50 layers with skip connections that allow information flows between early and later layers. We removed the last 2 fully connected layers of the ResNet-50 to extract the features from frames. The features extracted are of size  [2048, 7, 7], where 2048 is the number of channels and 7 is both the width and height. 

We report results using ResNet-50 pre-trained on Imagenet dataset~\cite{deng2009imagenet} and on Moments dataset~\cite{monfort2019moments}. The availability of ResNet-50's pre-trained versions on these big datasets is one of the main reasons why we considered it as our semantic feature extractor model. 

\paragraph{Predictive Network}
%structure of PN -- no of layers, how does it look like, main components and how does it run.
Predictive coding networks~\cite{lotter2016deep}, so called PredNet, was introduced for unsupervised feature learning, inspired by the brain and to leverage the availability of videos without annotations. It takes into account the time consistency by predicting the next frame, and propagating the prediction error. Figure~\ref{fig:prednet} illustrates the mechanics of a 2-layer PredNet.

In ~\cite{lotter2016deep}, the input to the PredNet is the image at the pixel level, and they used a 4-layer architecture. Instead, we use the features extracted from the top layers of the ResNet-50, which contain the semantic information necessary for performing action recognition. Thus, we predict and propagate the error at the feature level, instead of at the pixel level. 

The PredNet consists of four main components, which are R, E, A and \^{A} units, standing for representation, error, input and prediction, respectively. At time step 1, all R$^{1}_{0}$ and R$^{1}_{1}$ units are initialized as 0. Starting from the bottom layer 0, after convolution, R$^{1}_{0}$ produces \^{A}$^{1}_{0}$, which is the prediction unit with the same dimension as A$^{1}_{0}$, the feature extracted from ResNet-50. E$^{1}_{0}$ is the concatenation of the absolute values between the subtractions of \^{A}$^{1}_{0}$ and A$^{1}_{0}$, A$^{1}_{0}$ and \^{A}$^{1}_{0}$. After convolution, A$^{1}_{1}$ is produced by E$^{1}_{0}$ as the input for the next layer of the same time step. Similarly, at layer 1, R$^{1}_{1}$ generates \^{A}$^{1}_{1}$ via convolution and E$^{1}_{1}$ comes from the subtractions between \^{A}$^{1}_{1}$ and A$^{1}_{1}$, A$^{1}_{1}$ and \^{A}$^{1}_{1}$. This marks the ending of the bottom-up process. Beginning from layer 1 at time step 1, with a convolution LSTM (convLSTM), R$^{1}_{1}$ and E$^{1}_{1}$ initialize R$^{2}_{1}$ at layer 1 time step 2. R$^{2}_{1}$, with E$^{1}_{0}$ and R$^{1}_{0}$, outputs R$^{2}_{0}$ at layer 0 time step2 via convLSTM, which concludes the process of top-down. These bottom-up and top-down processes occur recurrently across the time domain.

Since ResNet-50 has a deep structure and the feature map extracted have width and height of 7,  we design the PredNet model to have 2 layers. We use 30 frames as input from each video, therefore, the number of time steps here is 30, and it is equivalent to 3 seconds of video. The number of channels used in our 2-layer PredNet is of 64 for each layer. We tried other number of channels, such as 128, 256, etc, but more channels did not give much edge for the classification performance, only slowing the training processes and taking up more memory.
%how are the two working together
\paragraph{Overall Architecture} Figure~\ref{fig:onecol} illustrates the complete layout of our model. At each time step, we have features from ResNet-50 with size [2048, 7, 7] and PredNet with size [64, 7, 7] from its 2 layers. Max-pool is applied on the features, and then they are concatenated as one with size of [2176, 1]. This concatenated feature serves as the input for a classification layer. The classification layer predicts a score vector for each time step. Finally, we average the score vectors of all the time steps to have a final prediction score. 

\section{Experimental Set-up}
In this section we detail the datasets used for the experiments, as well as the implementation.

\subsection{Datasets}

\paragraph{HMDB51~\cite{kuehne2011hmdb}} We report results on HMDB51, which consist of around 7,000 clips with annotations of 51 human action categories. It is divided into three splits, and each split contains training and testing sets. This dataset has at least 70 training and 30 test clips for each split, most of which are from movies. 

\paragraph{UCF101~\cite{soomro2012ucf101}} It is a video dataset with 13,320 videos for 101 human actions collected from YouTube, and it is a standard dataset for benchmarking action recognition models.

\paragraph{Moments in Time~\cite{monfort2019moments}} It is a large-scale dataset with over one million video clips with total 339 action labels annotated by humans. Each video is on average 3 seconds long. It is a challenging dataset, as it contains human, animals, objects and natural phenomena as subjects performing the actions. It is suggested to consider both top-5 and top-1 accuracies to measure model's perfomance due to dataset's complexity and scale. We used ResNet-50 pre-trained on this dataset~\cite{monfort2019moments} as a frame level feature extractor, and also report results on the other datasets after pre-trained PredNet on it. For efficiency and memory constraints, we used only the first 155 classes with 250 videos per class for pre-training PredNet in the experiment.

\subsection{Implementation details} 

\setlength{\tabcolsep}{4pt}
\begin{table*}
 \begin{center}
  \begin{tabular}{|c|c|c|c|c|c|c|}
   \hline
     \multicolumn{1}{|c|}{\textbf{Model Structure}} & \multicolumn{2}{c|}{\textbf{Pretrain on}} & \multicolumn{2}{c|}{\textbf{Finetune/Training}} & \multicolumn{1}{c|}{\textbf{Accuracy}} \\
    { } & {ResNet-50} & {PredNet} & {ResNet-50} & {PredNet} & { } \\
   \hline
	 ResNet-50 & ImageNet & - & classification layer  & - & 2.61\% \\
	 ResNet-50 & Moments & - & classification layer  & - & 55.03\% \\ 
     ResNet-50 + PredNet & ImageNet & - & fixed weight & from scratch & 13.66\% \\
     ResNet-50 + PredNet & Moments & - & fixed weight & from scratch & 58.43\% \\
	 ResNet-50 + PredNet & Moments & subset of Moments & fixed weight & finetune & 59.41\% \\
   \hline
  \end{tabular}
 \end{center}
\caption{Experiment result analysis of our model on the first split of HMDB51. ResNet-50 has two versions -- pre-trained on ImageNet or Moments. PredNet is either trained on HMDB51 split 1 from scratch or pre-trained on a subset of Moments then finetuned on the first split of HMDB51.}
\label{tab:tab1}
\end{table*}
\setlength{\tabcolsep}{2 pt}

%fps, data augmentation, preprocessing
\paragraph{Data preprocessing} Both Moments and HMDB51 datasets were processed to obtain videos of 30 frames per second. The majority of the videos in Moments and Hmdb51 have at least 90 frames. Hence, to make the most use of the video frames, we selected 90 consecutive frames randomly during both training and test phases. For those videos with less than 90 frames, the video clips are looped as many times as necessary. To reduce memory consumption, during training, 30 frames from the 90 are selected randomly according to the time sequence as input. During test time, 30 frames will be selected uniformly as input. Because PredNet needs to predict next time step's features continuously, each frame is center-cropped into width and height of 224 and normalized, following~\cite{paszke2017pytorch,TORCHVISION.MODELS}, to match the expected input of the pre-trained ResNet-50 model. This is done both at training and testing phases.   

\paragraph{Network parameters} We downloaded ResNet-50 pre-trained on ImageNet from the available Pytorch pre-trained models. We used the model pre-trained on Moments provided by the authors~\cite{monfort2019moments}~\footnote{Pre-trained model on Moments dataset available at: https://github.com/metalbubble/moments{\_}models}.

For training the deep networks, we use Standard stochastic gradient descent (SGD), with momentum set to 0.9 and weight decay to 0.001 in all cases. The batch size is 256 and the initial learning rate is 0.0064. As soon as the validation loss saturates, 10x reduction of the learning rate is applied. We train the PredNet with the action classification layers model for 40 plus epochs on HMDB51. For training or finetuning on Moments, there we use only 10 plus epochs due to time and memory constraints. Models are all implemented in Pytorch.

\section{Experiments}
We first perform an analysis of the components of our proposed model, and then we report results in comparison to state-of-the-art.

\subsection{Analysis of our Model}
We design several experiments to analyze each part of our model: 

\begin{enumerate}
    \item The last layer of ResNet-50 pre-trained on ImageNet dataset is finetuned on the first split of HMDB51.
    
    \item The last layer of ResNet-50 pre-trained on Moments dataset is finetuned on the first split of HMDB51.
    
    \item Fix the weight of ResNet-50 pre-trained on ImageNet dataset, train PredNet on the first split of HMDB51.
    
    \item Fix the weight of ResNet-50 pre-trained on Moments dataset, train PredNet on the first split of HMDB51.
    
    \item Fix the weight of ResNet-50 pre-trained on Moments dataset, finetune PredNet pre-trained on Moments on the first split of HMDB51.
    
\end{enumerate}

Experiments 1 and 2 are to compare the performance difference of ResNet-50 with different pre-trained weights, and without the temporal features. Experiments 3 and 4 are to exhibit how effective PredNet is with the two pre-trained ResNet-50 models, respectively. Experiment 5 pre-trains PredNet on part of Moments dataset, and is to verify whether PredNet pre-trained on Moments can increase classification accuracy with respect to training it from scratch.

 We report the results of the analysis on HMDB51 dataset in Table~\ref{tab:tab1}. We can see that ResNet-50 pre-trained on Moments dataset achieves a much higher accuracy than using the pre-trained model in ImageNet dataset, after finetuning the classification layer on HMDB51. The accuracy of the first one is of 2.62\% while of the other is 55.03\%. This indicates that using the features of a pre-trained model for object classification is not a good representation for action recognition. If we add PredNet on both scenarios, and train the PredNet from scrath on HMDB51, there is a significant boost in accuracy for both pre-trained ResNet-50, an 11\% increase for ImageNet pre-train ResNet, and more than 3\% increase for the Moments pre-trained one. This indicates that PredNet is able to capture addicitonal information from the video sequence, which we believe are related to motions, as PredNet propagates the prediction error of consecutive frames. 
 
 We also report results when pre-training the PredNet on part of a subset of Moments dataset (recall that for this we use only 155 classes and 250 videos per class), and use the fixed weights from the ResNet pre-trained on all of  Moments dataset. After pre-training PredNet on Moments, we can observe that there is a slight improvement of the accuracy prediction on HMBD51, achieving 59.41\%. We believe that this accuracy can be substantially improved if using all the Moments dataset videos available and train for more epochs, as well as using a deeper PredNet architecture. We haven't performed the experiment due to limitations to GPU and memory access. 

\begin{figure}[t]
\begin{center}
   \includegraphics[width=1\linewidth]{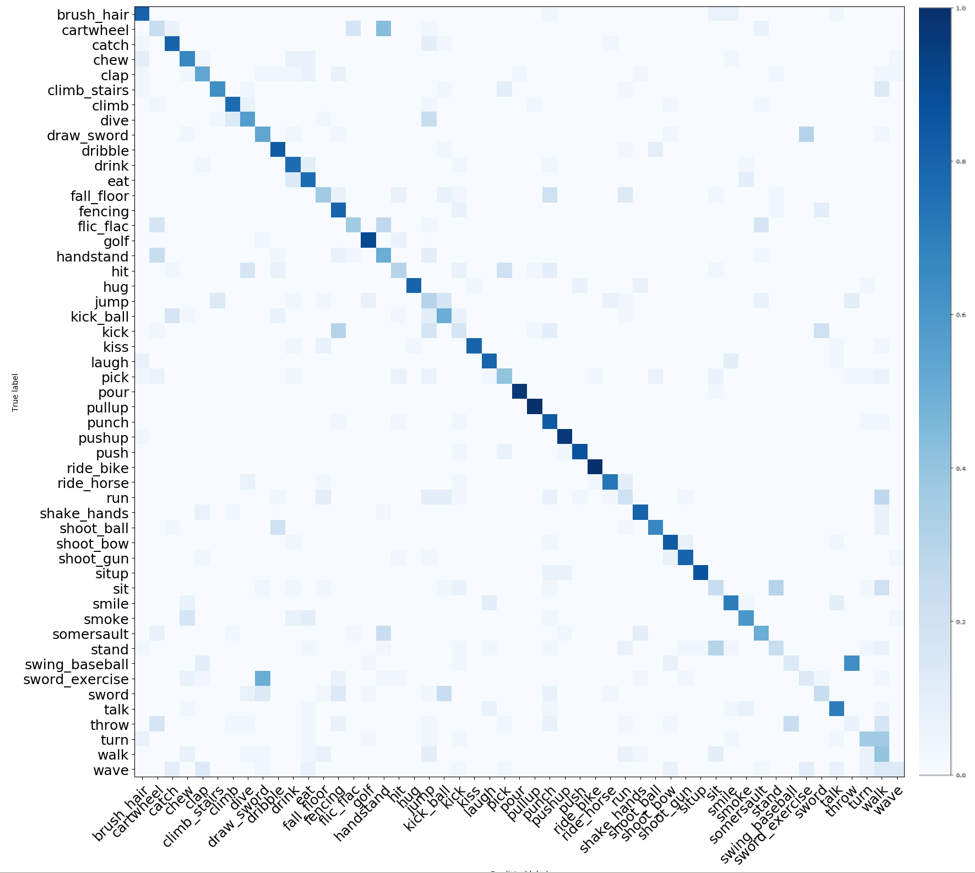}
\end{center}
   \caption{Normalized confusion matrix for ResNet-50 + PredNet on HMDB51, where ResNet-50 is pre-trained on Moments, PredNet on a subset of Moments then finetuned on HMDB51. The x-axis is prediction and the y-axis is the true label.}
\label{fig:conf}
\end{figure}

We compute the confusion matrix, as illustrated in Figure~\ref{fig:conf}. The results on HMDB51 are obtained with our best performing model including ResNet-50 and PredNet, both pre-trained on Moments dataset. We can see that many actions, such as pour, pullup, ride bike and push up, are correctly classified by the model. Yet, actions that are very similar between them, such as stand and sit, run and walk, are confused by the model. This could be because the number of frames in HMDB51 videos ranges between 1,000 and 17, and we always take a fix number of frames of 30. This might cause some information lost. Incorporating all the frames is computationally expensive, yet doable. 
Other actions that the model fail to recognize are those not included in Moments datasets, such as sword exercises and draw sword in HMDB51. For this 2 actions, the videos are extracted from movies, and the different actions are performed by the same actor in the same background, hence, it is indeed difficult to differentiate between them, even for human beings.

\setlength{\tabcolsep}{5pt}
\begin{table*}[t!]
\begin{center}
\label{table:results101}
\scalebox{0.97}{
					\begin{tabular}{lclclclcl}
						\hline
						\multicolumn{2}{c}{\textbf{UCF-101}}							&\multicolumn{2}{c}{\textbf{HMDB-51}} \\
						\hline
						Method							& Acc $\;\;$							& Method 								& Acc \\
						\hline
					
					    ImageNet Pretrained~\cite{simonyan2014two}& 73.0\%  & ImageNet Pretrained~\cite{simonyan2014two} & 40.5\% \\
						Spatial model~\cite{simonyan2014two}& 73.0\%  & Spatial model~\cite{simonyan2014two} & 40.5\% \\
						Richard and Gall~\cite{richardbow}						& 73.3\%			&Richard and Gall~\cite{richardbow} & 50.6\% \\
					
						iDT \cite{wang2013action}		& 85.9\%							& iDT \cite{wang2013action}   		& 57.2\% \\
						%Two-stream model~\cite{simonyan2014two}& \textbf{86.2}\%  	&Two-stream model~\cite{simonyan2014two} & 58.\% \\
						Dense Predictive Coding~\cite{Han19}	& 75.7\% & Dense Predictive Coding~\cite{Han19} 		& 35.7\% \\  					 %Action-Gons~\cite{wang2014action} 		& 58.9\% \\
						%CNN-$fc6$~\cite{zha2015exploiting}& 79.3\% 						& ~\cite{cordelia15} & 0.5\% \\
						%CNN-$fc6$+iDT+FV~\cite{zha2015exploiting}& 89.62\% 		& ~\cite{cordelia15} & 0.5\% \\			 	
						\hline
						Our Model 								&\textbf{82.69}\%  			& Our Model 						& \textbf{56.91}\% \\
					
						\hline
					\end{tabular}}
				\end{center}
					\caption{Comparison results on UCF 101 and HMDB51 action recognition datasets. }
			\end{table*}
			\setlength{\tabcolsep}{2.pt}

%\paragraph{Results on HMDB51} The accuracies for the three splits of HMDB51 are 59.41\%, 55.62\% and 55.69\% respectively. The average accuracy is 56.91\%.

\subsection{Comparison to Previous Methods}
We compare the results on HMDB51 and UCF101 datasets obtained with our proposed model, ResNet-50 and PredNet pre-trained on Moments dataset, to previous methods. 
%In this experiment, we compared the proposed approach with ImageNet Pretrained~\cite{simonyan2014two},iDT~\cite{wang2013action}, Richard and Gall~\cite{richardbow}, Spatial model~\cite{simonyan2014two} and Dense Predictive Coding ~\cite{Han19} following the standard evaluation of on HMDB51 and UCF101 datasets. Comparing our approach with the method proposed in Dense Predictive Coding~\cite{Han19} we put an emphasis on extracting motion information from 

\paragraph{Results on HMDB51}
We report results on HMDB51 of our model compared to previous models in Table~\ref{table:results101}. We observe that our model outperforms by a large margin several  state-of-the-art methods which, as our model are based on RGB frames only as input, namely~\cite{richardbow} and the spatial model of~\cite{simonyan2014two}. The two-stream model, based on convolutional neural networks  proposed by~\cite{simonyan2014two}, also includes optical flow explicitly and used as input to the system as an addition to the RGB frames. Their model achieves an accuracy performance, which is only slightly higher than the one of our model, that only has RGB frames as input. 

Please note that the accuracy of our model reported on this table is the average across 3 splits of HMDB51 to be able to compare to methods from the literature, while in Table~\ref{tab:tab1} is for one split only. 

\paragraph{Results on UCF101}
We also perform the evaluation on UCF101, which is a bigger dataset than HDMB51. A comparison between our method with state-of-the-art results is shown in~\ref{table:results101}.  Similar conclusions as with UCF-101 can be extracted. Our model gets 82.69\%, only using RGB frames as input,   outperforming the models using just spatial information, and is just slightly below the two stream model~\cite{simonyan2014two} which achieves 86.2\%.  

Simultaneously to ours, a similar framework for action recognition has been proposed  by Han \etal~\cite{Han2019predcod}, which also uses predictive coding networks. They report an accuracy of 75.7\% and 
35.7\% for UCF-101 and HMBG-51, respectively, which is significantly lower than ours. The main differences are in the architecture of the networks and the dataset used for pre-training the model, as they use Kinetics dataset~\cite{kay2017kinetics} and we use Moments in time. 

\section{Conclusions}
In this paper, we propose a model for action recognition that takes only RGB frames as input, and extracts motion features leveraging predicting coding networks (PredNets). We explored the combination of a spatial model ResNet-50 with PredNet for the task of action recognition on two classical datasets, HMDB51 and UCF101, using Moments datasets as a large dataset for pre-training the models. Our results demonstrate the power of PredNet as a model that uses only appearance features, and can extract motion information without computing optical flow. We believe that predictive coding networks have a great potential to exploit intrinsic information from unlabeled videos to learn meaningful representations, and can be applied to many other tasks besides action recognition, such as anomaly detection, navigation, and self-learning in robotic applications.

{\small
\bibliographystyle{ieee}
\bibliography{egbib}
}

\end{document}